\documentclass[lettersize,journal]{IEEEtran}
\usepackage{amsmath,amsfonts}
\usepackage{algorithmic}
\usepackage{array}
\usepackage[caption=false,font=normalsize,labelfont=sf,textfont=sf]{subfig}
\usepackage{textcomp}
\usepackage{stfloats}
\usepackage{url}
\usepackage{verbatim}
\usepackage{graphicx}
\def\BibTeX{{\rm B\kern-.05em{\sc i\kern-.025em b}\kern-.08em
    T\kern-.1667em\lower.7ex\hbox{E}\kern-.125emX}}
\usepackage{balance}
\usepackage{booktabs} 
\usepackage{pifont}
\newcommand{\cmark}{\ding{51}}%
\newcommand{\xmark}{\ding{55}}%
\usepackage{multirow}
\usepackage{siunitx}

\begin{document}
\bstctlcite{IEEEexample:BSTcontrol}
    \title{Thermal Detection of People with Mobility Restrictions for Barrier Reduction at Traffic Lights Controlled Intersections}
  \author{Xiao~Ni,
      Carsten~Kühnel,
      and~Xiaoyi~Jiang,~\IEEEmembership{Senior Member,~IEEE}

  \thanks{Manuscript received May 13, 2025. This work was funded in part by the "KIMONO-EF" Project by the Federal Ministry for Digital and Transport under Grant 45AVF3005A-E.}
  \thanks{Xiao Ni is with the Faculty of Mathematics and Computer Science, University of Münster, 48149 Münster, Germany (e-mail: xiao.ni@uni-muenster.de), and also with University of Applied Sciences Erfurt, 99085 Erfurt, Germany.}
  \thanks{Carsten Kühnel is with the Faculty of Business, Logistics and Transport, University of Applied Sciences Erfurt, 99085 Erfurt, Germany (e-mail: carsten.kuehnel@fh-erfurt.de).}%
  \thanks{Xiaoyi Jiang is with the Faculty of Mathematics and Computer Science, University of Münster, 48149 Münster, Germany (e-mail: xjiang@uni-muenster.de).}}


\maketitle

\begin{abstract}
Rapid advances in deep learning for computer vision have driven the adoption of RGB camera-based adaptive traffic light systems to improve traffic safety and pedestrian comfort. However, these systems often overlook the needs of people with mobility restrictions. Moreover, the use of RGB cameras presents significant challenges, including limited detection performance under adverse weather or low-visibility conditions, as well as heightened privacy concerns. To address these issues, we propose a fully automated, thermal detector-based traffic light system that dynamically adjusts signal durations for individuals with walking impairments or mobility burden and triggers the auditory signal for visually impaired individuals, thereby advancing towards barrier-free intersection for all users. To this end, we build the thermal dataset for people with mobility restrictions (TD4PWMR), designed to capture diverse pedestrian scenarios, particularly focusing on individuals with mobility aids or mobility burden under varying environmental conditions, such as different lighting, weather, and crowded urban settings. While thermal imaging offers advantages in terms of privacy and robustness to adverse conditions, it also introduces inherent hurdles for object detection due to its lack of color and fine texture details and generally lower resolution of thermal images. To overcome these limitations, we develop YOLO-Thermal, a novel variant of the YOLO architecture that integrates advanced feature extraction and attention mechanisms for enhanced detection accuracy and robustness in thermal imaging. Experiments demonstrate that the proposed thermal detector outperforms existing detectors, while the proposed traffic light system effectively enhances barrier-free intersection.  The source codes and dataset are available at https://github.com/leon2014dresden/YOLO-THERMAL.
\end{abstract}


%
\IEEEpeerreviewmaketitle


\section{Introduction}

\IEEEPARstart{T}{he} development of safe and accessible intelligent transportation system is crucial for promoting mobility and safety for all individuals \cite{Wang10247090, Sharma10101716, Roters5971774}, particularly those with disabilities \cite{ASIEDUAMPEM2024101935, Llorca7904784}. Urban intersections play a pivotal role in managing diverse traffic flows, including pedestrians, cyclists, drivers and public transportation users. However, conventional traffic systems are often inadequate in addressing the unique challenges faced by individuals with mobility restrictions due to fixed signal timings of traffic light and insufficient adaptive features \cite{Wunderlich4584232, Wang21093096}. Crossing streets poses a disproportionate risk for individuals with mobility restrictions. For instance, the mortality rate of wheelchair users is 36 percent higher in car-related accidents compared to standing pedestrians \cite{Kraemere008396}. This underscores the need for innovative approaches to develop barrier-free intersections that ensure safety, comfort, and accessibility for everyone.

A barrier-free intersection is designed to ensure universal accessibility, providing safe and convenient passage for all users, including pedestrians, cyclists, and individuals with mobility restrictions. Barriers to achieving such intersections can be categorized into four broad groups: (1) attitudinal-related barriers, (2) physical barriers, (3) policy- and program-related barriers, and (4) information and communications technology (ICT)-related barriers \cite{ASIEDUAMPEM2024101935}. This study focuses on the ICT-related category, where two of the most prevalent challenges are crossing signal times that are too short for people with mobility restrictions and the lack of automatically triggered auditory signals at crossings.

To address these challenges, this paper proposes a detector-based intersection control system capable of dynamically adjusting crossing signal times and providing auditory cues. When individuals with mobility restrictions cross the intersection, the system extends the crossing interval to ensure sufficient time for safe passage. Similarly, when visually impaired people attempt to cross the intersection, the system trigger the auditory signal to aid in orientation and guide them safely across the street. By tailoring signal durations and amplifying auditory guidance to accommodate individuals with mobility restrictions, the proposed system enhances safety and accessibility across urban road networks, particularly for people with mobility restrictions.

However, achieving barrier-free intersections with detectors is fraught with challenges. Conventional RGB cameras, used in various adaptive traffic systems, struggle in low-light and adverse weather conditions, limiting their reliability to detect pedestrians, especially at night or during harsh weather. Since these systems primarily depend on the RGB camera, the poor quality of the captured image under such conditions leads often to failures in key object detection tasks \cite{HUANG2008432, Liu2021ImageAdaptiveYF, Sindagi2020, KAUR2023103812}. Furthermore, privacy concerns associated with RGB camera systems discourage their deployment in sensitive urban spaces.

To overcome these limitations, we propose the incorporation of infrared sensors, which are merely affected by external illuminating and environmental conditions \cite{Gade2014}. Although thermal imaging offers robustness in adverse conditions, its unique characteristics, such as the lack of color and fine texture details, as well as difficulties in distinguishing objects with similar heat signatures, still pose significant hurdles for accurate detection \cite{BUSTOS2023126804}. In addition, while high-performance infrared cameras available for consumer applications are expensive, their resolution is still limited to a maximum of $640\times480$ pixels, significantly lower than RGB cameras, which can achieve resolutions of up to 4 megapixels.

To address these issues, we propose a lightweight real-time object detection framework built upon the YOLOv8 architecture, which is chosen for its optimal trade-off between speed and accuracy. This framework is specifically designed to tackle the inherent limitations of thermal imaging, such as low resolution, minimal texture details, and difficulties in distinguishing small objects. To achieve this, we integrate integrate the SPPFCSPC module \cite{Li2023}, the SPD-Conv module \cite{Sunkara2023}, and the Triplet-Attention mechanism \cite{Misra9423300}, which enhance feature extraction, object representation, and differentiation, improving overall detection performance.

For an accurate detection of individuals with various mobility restrictions in intersections, we construct a thermal dataset focused on people with disabilities. This thermal dataset comprises 11196 manually annotated images, capturing a diverse range of people with mobility aids, assistive devices, and other characteristics specific to the mobility restrictions.

This paper offers three primary contributions:
\begin{itemize}
    \item We build a specialized thermal infrared dataset focused on people with mobility restrictions, addressing critical gaps in existing datasets by incorporating previously underrepresented scenarios and environmental conditions.
    \item We propose a detector-based intersection control system that dynamically adjusts crossing signal times and increases the volume of auditory cues, specifically designed to improve accessibility for individuals with mobility and visual impairments.
    \item We propose a novel framework that addresses low resolution and limited texture in thermal imaging by integrating SPPFCSPC, SPD-Conv, and Triplet-Attention modules. Experiments validate its superior performance and robustness.
\end{itemize}
    
The remainder of this paper is structured as follows: Section \ref{Related work} provides a brief overview of thermal infrared datasets, datasets specifically focused on people with mobility restrictions, real-time object detection methodologies. Section \ref{TD4PWMR: Thermal Dataset for People with Mobility Restrictions}  introduces the proposed thermal dataset TD4PWMR, describing its collection process, annotation strategy, and dataset characteristics. Section \ref{Thermal Detector-based Intersection Control} presents a novel approach to intersection management that leverages thermal imaging-based pedestrian detection to dynamically adjust traffic lights, accommodating the specific needs of individuals with mobility restrictions. Section \ref{YOLO-Thermal} elaborates on the proposed thermal object detection framework, detailing its key components. Section \ref{Experiments} conducts a comprehensive performance evaluation, highlighting the effectiveness of our proposed model through extensive experiments. Finally, Section \ref{Conclusion} concludes the paper by summarizing key findings.

\section{Related work}
\label{Related work}

\subsection{Thermal Datasets}
There is currently no standardized and specialized thermal dataset for pedestrians in traffic environments. However, several existing datasets can be used to perform a simple evaluation of thermal detectors.

\begin{enumerate}
    \item OSU Color-Thermal: The OSU dataset \cite{DAVIS2007162} is a thermal and RGB image fusion dataset designed for object detection and tracking. It focuses exclusively on pedestrians and all six videos are captured with a low-resolution fixed thermal sensor ($320\times240$ pixels), the Raytheon PalmIR 250D, in static backgrounds.
    \item FLIR: The FLIR dataset \cite{SWAP} is specifically designed for research on the visible and thermal sensor fusion. It contains 9711 thermal images across 15 different object categories, captured from a Teledyne FLIR Tau thermal camera with a resolution of $640\times512$ pixels.
    \item KAIST: The KAIST dataset \cite{Choi8293689} integrates thermal imaging and other sensor modalities to address the challenges in autonomous and assisted driving for day and night. The dataset includes 8970 thermal images with the same resolution of $640\times480$ pixels, using a FLIR A655Sc thermal camera. However, The link provided in the paper for accessing the dataset and its associated toolkits is no longer functional.
    \item DENSE. The DENSE dataset \cite{Bijelic9157107} is a multimodal adverse weather that includes data from a variety of sensors, such as cameras, LiDAR, radar and infrared sensors. The dataset was collected over a distance of 10000 km during driving in Northern Europe, providing a diverse range of challenging environmental conditions. Thermal images were captured from an Axis Q1922 FIR camera, which offers a resolution of $640\times480$ pixels, in dynamic and moving backgrounds. In total, the dataset includes 11500 thermal images. Although the evaluation tools remain available, the dataset's website is no longer accessible, restricting users from obtaining the dataset.
\end{enumerate}

\begin{table}[tbp]
\setlength\tabcolsep{0pt} 
\caption{Comparison of the proposed thermal dataset TD4PWMR to existing thermal datasets.} 
\begin{tabular*}{\columnwidth}{@{\extracolsep{\fill}} ll *{5}{l} }
   \toprule
   \multicolumn {2}{l}{} &
   \multicolumn {1}{l}{OSU} & FLIR & KAIST & DENSE & Ours \\
   \midrule
           & Resolution   &  $320\times240$ & $640\times512$ & $640\times480$ & $640\times480$ & $640\times512$  \\ 
           & Bit Depth & 8 & 8 & 14 & 8 & 8 \\
           & Total Frames & 8544 & 9711 &  8970 & 11500 &  11196  \\
           & Availability & \cmark & \cmark & \xmark & \xmark &  \cmark \\
   \bottomrule
\end{tabular*}
\label{Comparison of thermal datasets}
\end{table}

Table \ref{Comparison of thermal datasets} provides a comparative overview of the proposed TD4PWMR dataset and existing thermal datasets based on various characteristics such as resolution, bit depth, total frames, and availability.

\subsection{Datasets for People with Mobility Restrictions}
Detecting people with mobility restrictions is a critical aspect of building barrier-free systems. Several research efforts have focused on datasets to advance this domain.

The Mobility Aid dataset \cite{vasquez17ecmr, kollmitz19ras} is a specialized people with mobility restrictions in indoor environments, such as hospitals and public buildings. This dataset includes over 17,000 annotated RGB-D images, emphasizing privacy by leveraging depth data. This approach supports real-world applications where RGB camera might be limited due to privacy concerns.
This dataset includes five distinct classes: pedestrians, person in wheelchairs, pedestrian pushing a person in wheelchairs, person with crutches, and person using a walking frame. This dataset represents significant steps in addressing the challenges of detecting people with disabilities in structured environments. However, It still remains constrained to indoor scenarios, such as hospitals or controlled laboratory settings, limiting their applicability in outdoor or unstructured environments.

In \cite{Mukhtar8634731}, a custom dataset is built for disabled people with mobility aids like wheelchair, crutch, walking frame, walking stick, and mobility scooter. This dataset was manually annotated and contains 5,819 images. In this work, pedestrians and mobility aids are detected separately, and their association is determined based on their spatial proximity.

\subsection{ICT-related Approaches for Barrier-free Intersection}
Ensuring accessibility at urban intersections is critical to creating inclusive cities for individuals with mobility restrictions. This section will be devoted to describing research on ICT-related approaches to reducing barriers at intersections.

ICT-related barriers often pose more challenges for people whose disabilities affect hearing, speaking, reading, and who rely on alternative forms of communication \cite{ASIEDUAMPEM2024101935}. Recent efforts have introduced smartphone-based accessibility solutions. For visually impaired people, 'smart' crosswalk systems have been designed with Bluetooth beacons that communicate with a custom smartphone app \cite{bustos2022improving}.  By measuring the strength of the received signal, these systems approximate the location of a user in relation to the crosswalk and provide audio or haptic guidance for safer navigation. However, many such applications require users to interact with complex interfaces, posing usability problems and limiting their reach \cite{Stefanov2022}.

An alternative approach proposed in \cite{Llorca7904784} integrates stereo vision–based object detection with active Bluetooth beacons to localize people with disabilities in real time. In this system, active Bluetooth beacons not only support localization but also send disability-related information to the infrastructure, enabling the traffic light to deliver adaptive responses tailored to different impairment types. Nevertheless, this approach relies on people with disabilities obtaining and consistently carrying a dedicated disability identification device issued by local authorities, which may pose challenges in terms of convenience.

Inspired by these efforts, our solution incorporates thermal camera–based detection to dynamically adjust traffic signal durations for individuals with mobility restrictions and amplify auditory signals for visually impaired pedestrians. Unlike prior approaches, which rely on smartphones or tags, our solution requires no personal devices, thereby offering a privacy-preserving, and universally accessible framework for creating barrier-free intersections.

\subsection{Real-time Object Detectors}
YOLOv1 \cite{Redmon7780460} marked a paradigm shift in object detection by introducing the first CNN-based one-stage detector capable of achieving real-time performance. Over the years, the YOLO family has undergone significant improvement, consistently exceeding other one-stage detectors \cite{Lin8417976, Liu46448} and establishing itself as the synonym for real-time object detection \cite{Lv2023DETRsBY}. YOLO detectors are broadly categorized into two types: anchor-based \cite{bochkovskiy2020, Wang9577489, Redmon2016, redmon2018, Wang10204762} and anchor-free methods \cite{Ge2021, Li2023}. Both categories effectively balance accuracy and speed, making them versatile tools for a wide array of practical applications.

In recent years, transformer-based detectors have attracted increasing attention. DETR \cite{carion2020end}, the first end-to-end detector built on Transformer, removes both hand-crafted anchors and non-maximum suppression. Although DETR offers clear advantages, it also faces challenges such as slow training convergence and high computational cost. Current DETR variants remain too compute-intensive for real-time detection. To address these limitations, RT-DETR \cite{Lv2023DETRsBY} seeks to reduce computational cost and optimize query initialization, aiming to achieve performance comparable to real-time object detectors.

There are several specialized detectors designed for thermal images. However, \cite{Shyam10483657, Jiang25060914} are optimized for extremely low-resolution thermal images, such as $160\times120$ pixels, and rely on super-resolution techniques to enhance image quality. This approach is not suitable for our use case, as our thermal camera operates at a relatively high resolution compared to those scenarios. In \cite{Kri9133581}, YOLOv3 model has been directly implemented for thermal object detection by retraining it on a thermal dataset. However, this approach does not adapt the model to the unique characteristics of thermal images, limiting its ability to fully leverage the potential of thermal data. Additionally, these papers of existing thermal detectors do not provide publicly available code implementations. For the reasons outlined above, we exclude them from the scope of this paper.

\begin{figure*}[htbp]
    \centering
    \captionsetup[subfloat]{labelformat=empty, justification=centering, font=scriptsize}

    \subfloat[person without mobility restrictions]{%
        \includegraphics[width=0.16\textwidth]{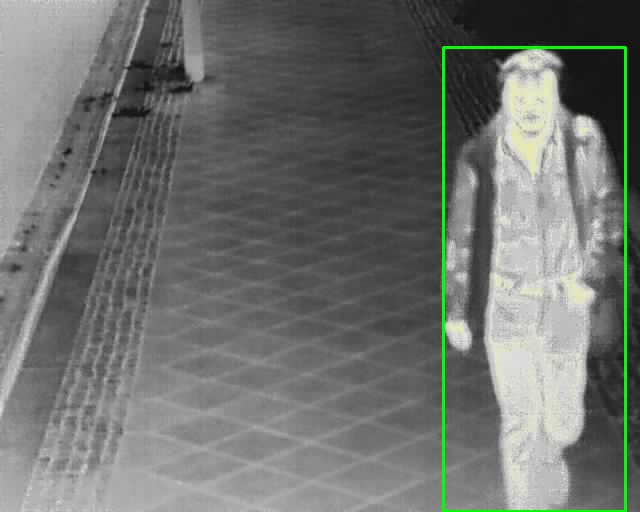}%
    }
    \hfill
    \subfloat[person with wheelchair]{%
        \includegraphics[width=0.16\textwidth]{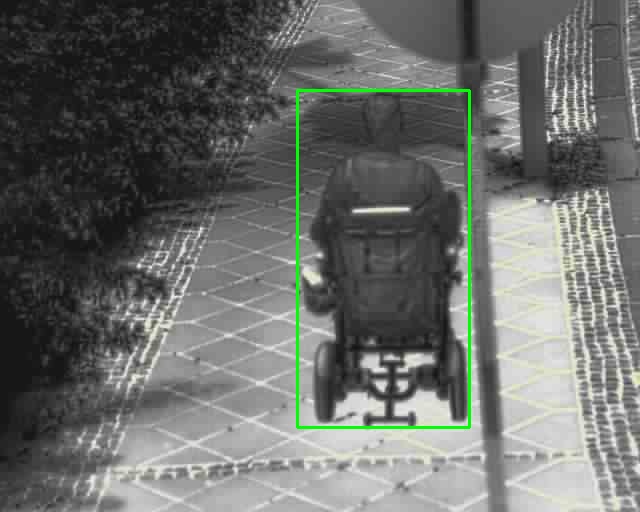}%
    }
    \hfill
    \subfloat[person with rollator]{%
        \includegraphics[width=0.16\textwidth]{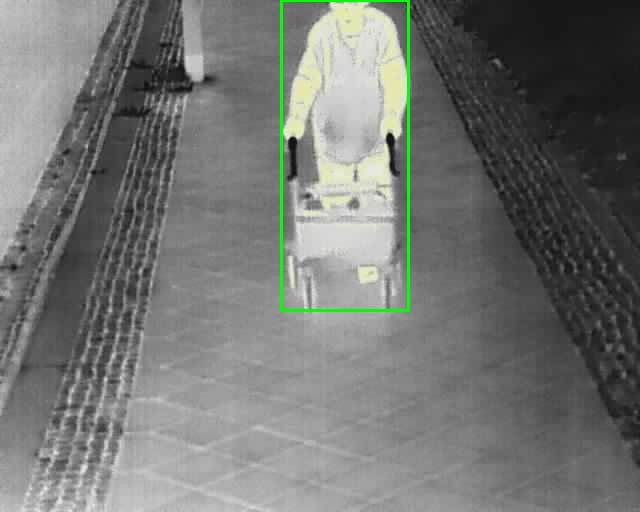}%
    }
    \hfill
    \subfloat[person with crutches]{%
        \includegraphics[width=0.16\textwidth]{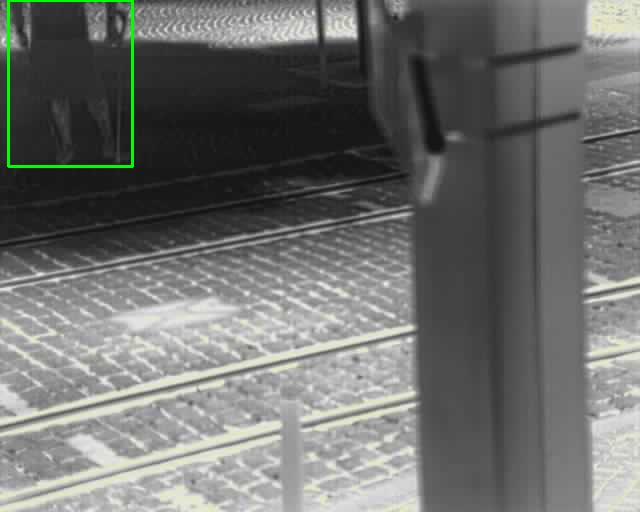}%
    }
    \hfill
    \subfloat[person with blindstick]{%
        \includegraphics[width=0.16\textwidth]{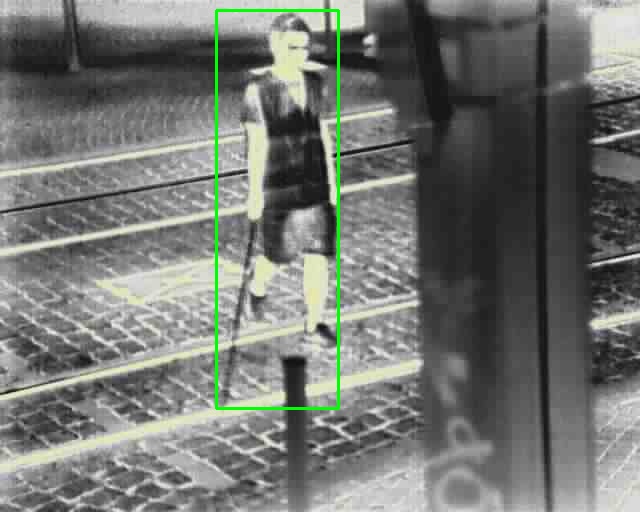}%
    }
    \hfill
    \subfloat[person with luggage]{%
        \includegraphics[width=0.16\textwidth]{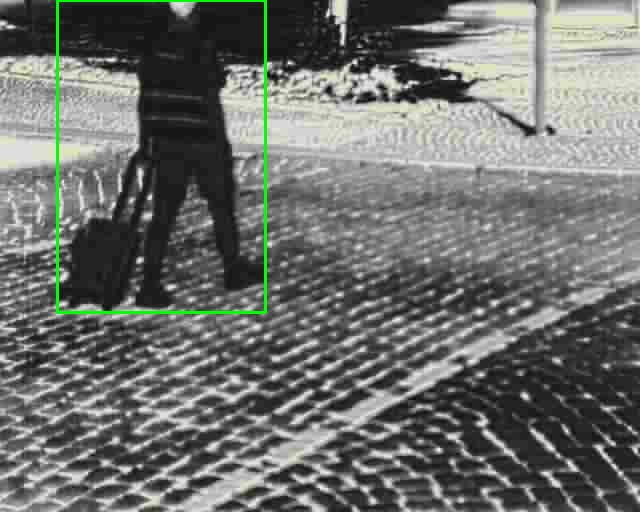}%
    }


    \subfloat[person with stroller]{%
        \includegraphics[width=0.16\textwidth]{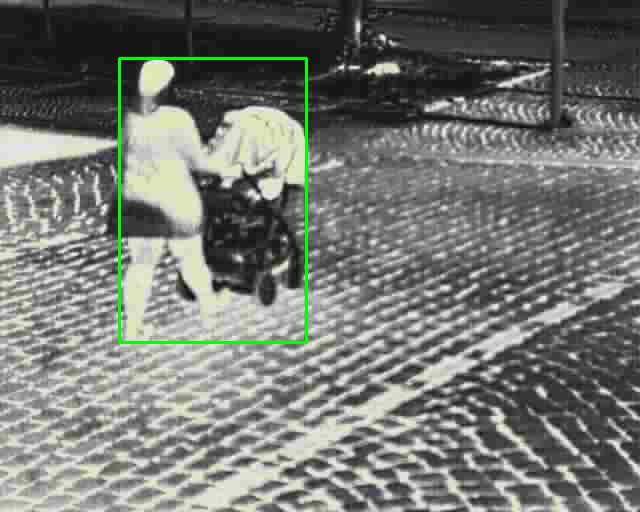}%
    }
    \hfill
    \subfloat[person with bicycle]{%
        \includegraphics[width=0.16\textwidth]{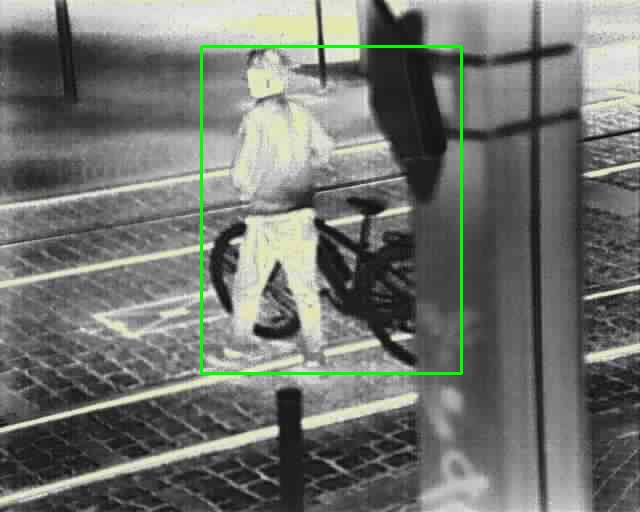}%
    }
    \hfill
    \subfloat[person with luggage trolley]{%
        \includegraphics[width=0.16\textwidth]{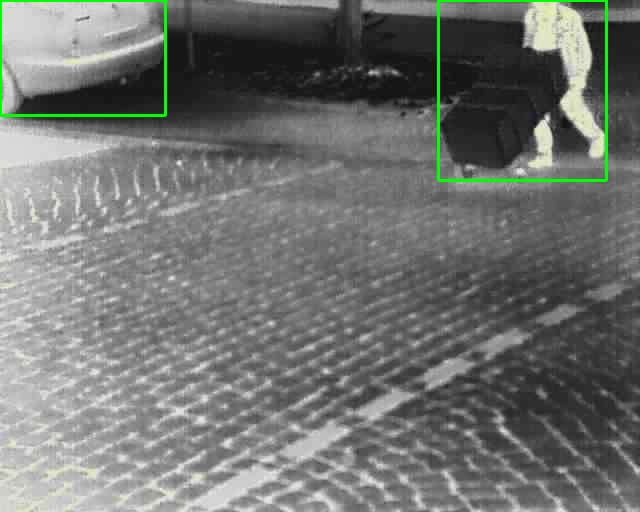}%
    }
    \hfill
    \subfloat[person with dog]{%
        \includegraphics[width=0.16\textwidth]{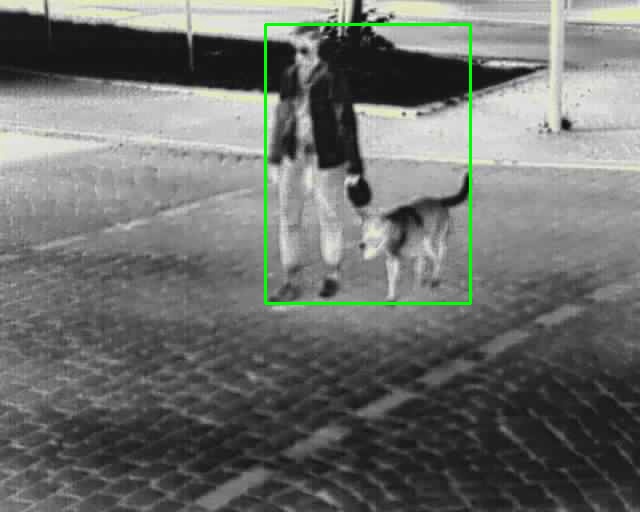}%
    }
    \hfill
    \subfloat[person with walking stick]{%
        \includegraphics[width=0.16\textwidth]{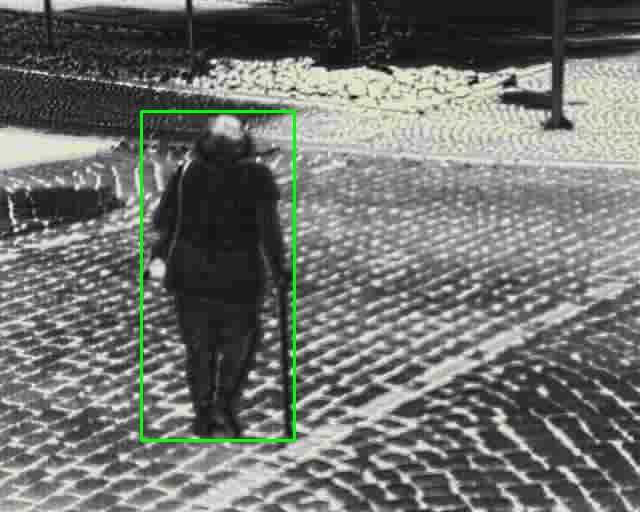}%
    }
    \hfill
    \subfloat[car]{%
        \includegraphics[width=0.16\textwidth]{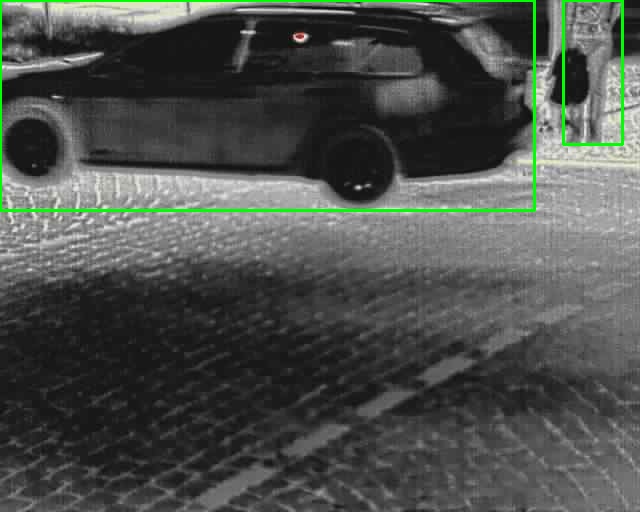}%
    }

    \caption{One representative example from each class in the proposed TD4PWMR dataset.}
    \label{fig:6x2_grid}
\end{figure*}

\section{TD4PWMR: Thermal Dataset for People with Mobility Restrictions}
\label{TD4PWMR: Thermal Dataset for People with Mobility Restrictions}
To develop barrier-free intersections while protecting the privacy of pedestrians, it is crucial to establish a comprehensive thermal dataset that specifically focuses on pedestrians with mobility restrictions.

We cover various areas at intersections, including sidewalks, waiting zones, and intersection crossings, capturing each location during different time slots (sunrise, morning, afternoon, sunset, night, and dawn). Since the thermal contrast ratio is strongly influenced by ambient heating conditions, seasonal variations significantly affect the thermal signature of the scenes. To account for this, we captured images across all four seasons, ensuring a comprehensive dataset that reflects diverse environmental and thermal conditions. Furthermore, we deliberately balanced the number of images per season, aiming for an approximately equal distribution to minimize seasonal bias during model training and evaluation. Specifically, the dataset includes 3,086 images captured in spring, 2,584 in summer, 2,534 in autumn, and 2,992 in winter.

Our recording sensors are mounted on intersection poles, with a NVIDIA Jetson AGX Orin housed within the traffic control cabinet. To ensure stable and high-speed data collection, we utilized a solid-state drive (SSD), which offers significantly greater bandwidth compared to a traditional hard disk drive (HDD). In addition, we employ CAT7 cables between the edge computing device and the network switch, which connects to each sensor via CAT6 cables and aggregates the collected data. This setup reduces transfer latency and increases bandwidth. A total of eight infrared cameras (FLIR ThermiBot2) have been installed at the intersection. The camera operates within a spectral range of 7.5–13.5 ${\mu}m$ with a native resolution of $640\times512$ pixels and supports focal lengths ranging from 9 $mm$ to 35 $mm$ for versatile deployment scenarios. The infrared camera is barely affected by changes in illumination, making it a reliable choice for operation in adverse weather conditions.  

To facilitate a variety of perception tasks, we manually annotated objects across all thermal video sequences. Prior to annotation, we established clear definitions for annotation targets and types. During the annotation process, the bounding boxes were drawn and the annotators assigned labels to objects by selecting from the following 12 predefined classes: person without mobility restrictions, person with wheelchair, person with rollator, person with crutches, person with blindstick, person with luggage, person with stroller, person with bicycle, person with shopping/luggage trolley, person with dog, person with walking stick, and car. For individuals assisted by mobility aids or restrained by mobility burden, the bounding box is drawn to encompass both the person and the associated object, treating them as a single object. Since these mobility aids and mobility burden are in constant use and often become partially occluded by the user’s own body, making detection more challenging. By combining the person and the associated item into a single class, we leverage their relative positioning to enhance object detection accuracy and robustness. Figure \ref{fig:6x2_grid} presents representative examples from each class, demonstrating the way we annotate the bounding boxes. In the images, facial features and specific identities remain difficult to discern due to the nature of thermal imaging. This characteristic ensures that surveillance and monitoring systems can operate effectively without compromising personal privacy.

To facilitate effective model training and evaluation, the dataset is divided into training (80\%) and evaluation (20\%) subsets. The training subset is used to train detection models, ensuring that they learn to accurately identify and classify people with mobility restrictions across varied scenarios. The evaluation subset is reserved for testing, providing an unbiased measure of the model's performance in detecting and classifying people with mobility restrictions in unseen conditions.

\begin{figure}[htbp]
  \includegraphics[width=3.5in]{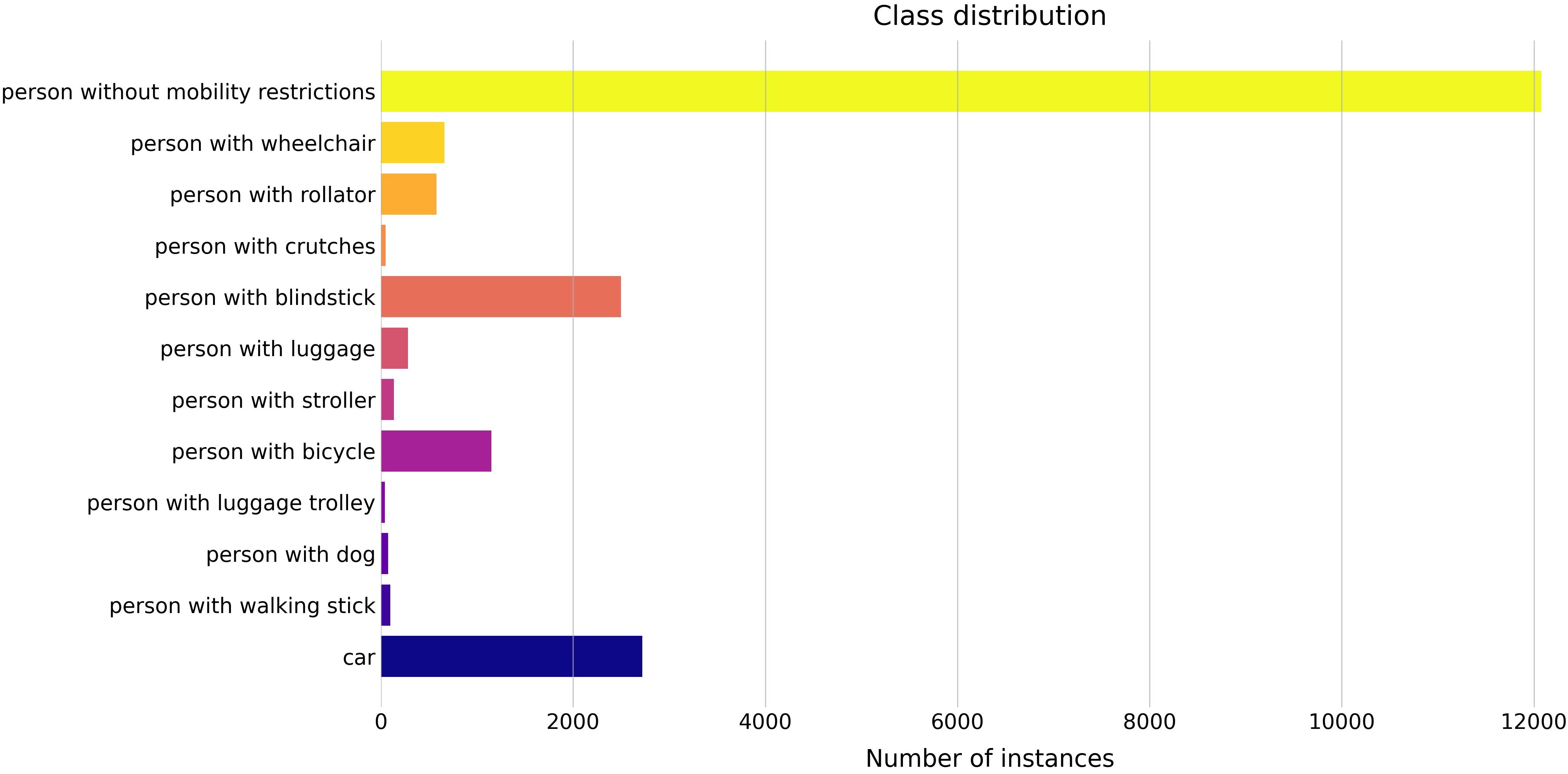}
  \\
  \caption{Class distribution, highlighting the significant imbalance in the frequency of different classes.}\label{dataset_labels}
\end{figure}

Figure \ref{dataset_labels} reveals that the dataset exhibits a significant class imbalance, as the frequency of people with mobility restrictions is much lower compared to that of people without mobility restrictions. This imbalance poses a substantial challenge for training detection models, as they tend to focus disproportionately on the majority classes, leading to insufficient learning for the minority classes. Consequently, rare classes, such as people with various assistive devices, may not receive adequate training, resulting in poor detection performance and reduced model generalization for these underrepresented categories.

\begin{table*}[htbp]
\centering
\caption{Categorization of pedestrians with mobility restrictions and their corresponding adaptation strategies of traffic light controller.}
\begin{tabular}{|l|l|p{10cm}|}
\hline
\textbf{Classes} & \textbf{Group} & \textbf{Strategy} \\
\hline
\multicolumn{1}{|l|}{\multirow{1}{*}{Person with wheelchair}} & \multirow{4}{*}{Person with walking impairments} & \multirow{2}{*}{Extend the green time $T_{g}$ of the current traffic direction,} \\
\multicolumn{1}{|l|}{\multirow{1}{*}{Person with crutches}} &  & \multirow{2}{*}{\textbf{if} person with walking impairments exists in the corresponding crossing area.}  \\
\multicolumn{1}{|l|}{\multirow{1}{*}{Person with rollator}} &  & \multirow{2}{*}{The maximal extension of the green time $T^{max}_{g, ext} = 6\,s$.} \\
\multicolumn{1}{|l|}{\multirow{1}{*}{Person with walking stick}} &  & \\
\hline
\multicolumn{1}{|l|}{\multirow{4}{*}{Person with blindstick}} & \multirow{4}{*}{Person with visual impairments} & Extend the green time $T_{g}$ of the current traffic direction and increase the volume of audible guidance, \\
\multicolumn{1}{|l|}{} &  & \textbf{if} person with visual impairments exists in the corresponding crossing area. \\
\multicolumn{1}{|l|}{} &  & The maximal extension of the green time $T^{max}_{g, ext} = 8\,s$. \\
\hline

\multicolumn{1}{|l|}{Person with luggage} & \multirow{5}{*}{Person with mobility burden} & \multirow{3}{*}{Extend the green time $T_{g}$ of 
the current traffic direction,} \\
\multicolumn{1}{|l|}{Person with stroller} &  & \multirow{3}{*}{\textbf{if} person with walking impairment exists in the corresponding crossing area.} \\
\multicolumn{1}{|l|}{Person with bicycle} &  & \multirow{3}{*}{The maximal extension of the green time $T^{max}_{g, ext} = 3\,s$.} \\
\multicolumn{1}{|l|}{Person with luggage trolley} &  &  \\
\multicolumn{1}{|l|}{Person with dog} &  &  \\
\hline
\end{tabular}
\label{classes_groups_strategies}
\end{table*}

\section{Detector-based smart traffic light for people with mobility restrictions}
\label{Thermal Detector-based Intersection Control}
The traffic light controller is equipped with two cameras at each crossing area, one for each direction of pedestrian movement. These cameras are mounted atop the traffic light poles to provide a wide field of view and ensuring comprehensive coverage of the crosswalk. In the case of a four-way intersection, this configuration results in a total of eight cameras. 

Table \ref{classes_groups_strategies} categorizes pedestrians with mobility restrictions into three primary groups: people with walking impairments, people with visual impairments, and people with mobility burden. The first group includes individuals who use assistive devices such as wheelchairs, crutches, rollators and walking sticks. The second group consists of visually impaired pedestrians, identified by the use of blind sticks. The third group comprises individuals with mobility burden, such as those carrying luggage, pushing strollers, bicycle and luggage trolleys, or walking with dogs. Each class is assigned a corresponding group based on their needs and challenges at crosswalks.

To enhance intersection accessibility and safety, the traffic light controller applies tailored adaptation strategies for each group. The 95\% design principle, widely used in ergonomic and infrastructure design, ensures accommodations for most users while preventing excessive customization needs \cite{Pheasant}. Our measurements indicate that the sufficient crossing time for 95\% of pedestrians with walking impairments is 5.8 seconds longer than that of the general population. For 95\% of visually impaired pedestrians, crossing takes an average of 7.9 seconds longer, while 95\% of pedestrians with mobility burden require an additional 2.7 seconds. Due to hardware constraints, traffic light controllers typically operate on integer-second timing intervals. As a result, green time extensions are rounded to the nearest whole second to ensure compatibility with existing infrastructure and to maintain synchronization across traffic control systems. Accordingly, the traffic light extends the green time by up to 6 seconds for individuals with walking impairments upon detection in the crossing area. When visually impaired pedestrians are detected in the crossing area, the maximum green phase extension is increased to 8 seconds, accompanied by an audible signals to aid navigation. Similarly, pedestrians with mobility burden trigger a maximum green time extension of 3 seconds if they are detected in the crossing area. These adaptive strategies enhance safety and accessibility of the intersection while minimizing disruptions to overall traffic flow.

When multiple pedestrians with mobility restrictions from different categories are detected in the crosswalk simultaneously, the traffic light controller prioritizes the adaptation strategy of the category with the highest priority. Based on an evaluation of mobility capability and safety concerns, the priority order is as follows: people with visual impairments receive the highest priority, followed by people with walking impairments, and finally, people with mobility burden.

This prioritization is driven by the level of vulnerability and the ability of each group to navigate intersections safely. Visually impaired pedestrians face the greatest challenges, as they rely primarily on auditory cues and tactile guidance, making it difficult to judge traffic conditions, detect signal changes, or react quickly to unexpected obstacles. Their increased reliance on external assistance justifies the longest possible green light extension and enhanced audible guidance signals to ensure they cross safely. People with walking impairments face significant crossing challenges due to reduced movement speed and increased physical effort required to navigate curbs and uneven surfaces. However, they can still rely on visual cues and have some degree of maneuverability, making them less vulnerable than visually impaired pedestrians. Consequently, they receive a moderate extension of green time to accommodate their slower pace while ensuring smooth traffic flow. Finally, pedestrians with mobility restrictions move at a slightly reduced pace but do not face significant physical or sensory barriers as the other two groups. Thus, they receive the shortest green time extension to minimize traffic disruption while still allowing additional crossing time when necessary.

To detect pedestrians with mobility restrictions, we employ an efficient detection algorithm. When the detector identifies a pedestrian belonging to a target group, it classifies the individual and generates a bounding box around them. We define the middle bottom point of the bounding box as the pedestrian's position. The crosswalk area is predefined by outlining a polygon-shaped region on the street map, and the system checks whether the position of the detected pedestrian falls within this designated area. This verification determines whether a person with mobility restrictions is present in the crosswalk, allowing the system to trigger appropriate adaptation strategies.

To increase the robustness of the traffic light control, we establish a reliable detection mechanism to determine the presence of pedestrians with mobility restrictions in the crosswalk. Once pedestrians with mobility restrictions are detected for the first time during the green phase, they are considered existing in the crossing area. However, accurately determining when they have fully exited the crosswalk remains a challenge. Due to occlusions and detection failures, the system can mistakenly conclude that they have already crossed when they are still in the intersection.

One potential solution to this issue is tracking-by-detection, which maintains pedestrian identities across frames and it could help to determine whether they have exited the crosswalk. However, this approach is not feasible in our system due to computational constraints. The detection algorithm is deployed on an edge computing device, the NVIDIA Jetson Orin, which has limited processing power. Given that the intersection is equipped with eight cameras, the system must process multiple video streams simultaneously. As a result, the interval between two processed frames for each camera is too large, making it impractical to apply tracking-by-detection algorithms that rely on high frame rate consistency.

To mitigate this, we implement a multi-frame validation approach, where the system only confirms the absence of pedestrians with mobility restrictions if they remain undetected for N consecutive frames after their initial detection in the crosswalk. This helps prevent premature termination of the extended green phase, ensuring sufficient crossing time. However, this approach also introduces a trade-off: If pedestrians with mobility restrictions have already exited the crosswalk and the system continues extending the green phase for the next $N$ frames, potentially reducing traffic efficiency. Optimizing the parameter $N$ is crucial to balancing robust protecting mechanism and smooth traffic flow, which we further analyze in section \ref{Experiments}.

\section
{YOLO-Thermal}
\label{YOLO-Thermal}
Although thermal sensors provide distinct advantages over RGB cameras, they are also known for their limitations, such as the absence of color and fine texture details, lower resolution, and difficulty in distinguishing between objects with similar heat signatures. These limitations render conventional RGB-based detection methods less effective for deployment with thermal cameras. In this paper, we introduce a novel detection model built upon the YOLOv8 architecture \cite{yolov82023}, which is selected for its good balance between accuracy and speed. The model is specifically tailored for thermal imaging by incorporating several new modules.

\subsection{Triplet-Attention}
Triplet Attention \cite{Misra9423300} is an attention mechanism designed to enhance feature representation in deep learning models by capturing cross-dimensional interactions. Unlike conventional attention modules that focus on either spatial or channel dependencies separately, Triplet-Attention integrates both by employing a three-branch structure.

Given an input tensor $X$ of dimensions $C \times H \times W$, the module builds dependencies across spatial $(H, W)$ and channel $(C)$ dimensions through three key transformations:
\subsubsection{Cross-Dimension Interaction}
Traditional attention mechanisms like SENet \cite{Hu8578843} and CBAM \cite{Woo1007} focus on independent spatial and channel-wise attention, potentially losing inter-dependencies. Conversely, Triplet Attention introduces a cross-dimension interaction strategy, ensuring that both spatial and channel dependencies are captured. The first branch models interactions between height $H$ and channel $C$ by permuting the input tensor, applying Z-Pool, and processing it through a convolutional layer. The second branch captures dependencies between width $W$ and channel $C$ similarly. The third branch follows CBAM’s spatial attention paradigm, focusing on $H-W$ dependencies.

\subsubsection{Z-Pooling}
Instead of direct global average pooling (GAP), Triplet Attention employs a novel Z-Pool operation that concatenates both global max pooling (GMP) and GAP along a particular dimension:
\begin{equation}\label{Z_Pooling}
    Z-pool(X)=[\text{MaxPool}_{0d}(X), \text{AvgPool}_{0d}(X)],
\end{equation}
where $0d$ denotes the dimension across which Z pooling operation is applied.

\subsubsection{Final Attention Weight Aggregation}
The refined feature representations from all three branches are aggregated as:
\begin{equation}\label{Aggregation}
    Y = \frac{1}{3}(\overline{X_1  W_1} + \overline{X_2  W_2} + X  W_3) \\
    = \frac{1}{3}(\overline{Y1} + \overline{Y2} + Y3),
\end{equation}
where $w_i$ denotes the attention weights computed in each branch and $\overline{Y_i}$ represents the $\ang{90}$ clockwise rotation applied to preserve the original input shape of $C \times H \times W$. 

To maximize its effectiveness in the YOLO architecture, Triplet-Attention module should be strategically placed immediately after the backbone of YOLO, specifically the custom CSPDarknet53. The backbone is responsible for extracting low-level to mid-level features from the input image, forming the foundation for subsequent detection stages. By integrating the triplet-attention module at this point, the model achieves improved localization and classification performance in thermal images, effectively mitigating the challenges posed by low-resolution and ambiguous thermal object boundaries.

\subsection{SPD-Conv}
SPD-Conv, or Space-to-Depth Convolution \cite{Sunkara2023}, is a building block designed to replace traditional strided convolution and pooling operations in convolutional neural networks (CNN). It is particularly effective for tasks involving low-resolution images and small object detection, where preserving fine-grained details is crucial.

In our proposed thermal dataset, the image resolution is limited to $640\times512$ pixels, and because the dataset is collected in outdoor environments, pedestrians can often appear far from the camera, resulting in small representations within the images. These two characteristics (low resolution and small object size) pose significant challenges for traditional down-sampling methods, which typically discard valuable fine-grained details.

SPD-Conv addresses these challenges by retaining fine-grained details during the space-to-depth transformation, redistributing them into the channel dimension for enhanced feature representation. This ensures that even subtle features of small objects, such as distant pedestrians in thermal images, are effectively captured and processed. As a result, SPD-Conv is particularly well-suited for thermal datasets with these demanding characteristics, offering superior performance compared to conventional approaches.

The module consists of two primary components: a Space-to-Depth (SPD) transformation followed by a non-strided convolution layer.

\subsubsection{Space-to-Depth Transformation}
Let $X$ be the input feature map with dimensions $S \times S \times C_1$, where $S$ is the size of length and width, and $C_1$ is the number of channels. The SPD transformation slices $X$ into submaps $f_{x,y}$ such that:
\begin{equation}\label{spd}
    f_{x,y} = X \left[x:S:scale, y:S:scale\right],
\end{equation}
where $scale$ is the downsampling factor.

This slicing generates $scale^2$ submaps, each of size $\frac{S}{scale} \times \frac{S}{scale} \times C_1$. These submaps are concatenated along the channel dimension, resulting in a transformed feature map $X^{\prime}$ with dimensions:
\begin{equation}\label{spd_dimension}
    X^{\prime} \in \mathbb{R}^{\frac{S}{scale} \times \frac{S}{scale} \times (scale^2 \cdot C_1)}.
\end{equation}

\subsubsection{Non-Strided Convolution}
After SPD transformation, a non-strided (i.e., $stride=1$) convolution layer with $C_2$ filters is applied, where $C_2 < scale^2 \cdot C_1$. This operation reduces the number of channels while preserving as much discriminative feature information as possible.
\begin{equation}\label{Non_Strided_Convolution}
    X^{\prime \prime} = Conv(X^{\prime}),
\end{equation}
where $ X^{\prime \prime} \in \mathbb{R}^{\frac{S}{scale} \times \frac{S}{scale} \times C_2} $.

We replace all convolutional layers in the YOLO model with SPD-Conv layers. By leveraging space-to-depth transformation, SPD-Conv effectively captures fine-grained features. Furthermore, this replacement helps mitigate the limitations of low-resolution thermal imaging, improving the model’s ability to detect and distinguish objects in thermal images.

\subsection{SPPFCSPC}
The Spatial Pyramid Pooling-Fast Cross Stage Partial Connections (SPPFCSPC) block \cite{Li2023} is a feature extraction component commonly utilized in deep learning models, particularly for object detection tasks. It combines Spatial Pyramid Pooling (SPP) \cite{He7005506}, which captures multi-scale spatial features by applying pooling operations at various scales, with Cross Stage Partial Connections (CSPC) \cite{Wang9150780}, which improves gradient flow and reduces computational redundancy by splitting feature maps into transformed and shortcut paths. This integration enables the module to extract both local and global contextual information effectively while maintaining lightweight computational requirements. By doing so, it significantly improves the ability to differentiate between objects in thermal images, which is critical for accurately identifying objects in scenes with sparse texture details - a common challenge in imagery captured by thermal cameras. SPPFCSPC is particularly well-suited for handling objects of varying scales and optimizing feature reuse, making it ideal for real-time applications and scenarios with limited computational resources.

We replace SPPF (Spatial Pyramid Pooling Fast) with SPPFCSPC because SPPFCSPC integrates CSP connections into the SPPF block, improving gradient flow and computation efficiency. This typically leads to higher accuracy in object detection while maintaining competitive speed.

\subsection{Quality Focal Loss}
Additionally, there is still another issue that comes from our proposed thermal dataset: class imbalance. Traditional methods to address class imbalance include re-sampling, which involves oversampling the minority class or undersampling the majority class. However, both approaches come with significant drawbacks. Oversampling can lead to overfitting, while undersampling reduces available training data \cite{Johnson}. A more effective solution is Focal Loss \cite{Lin8417976}, which adaptively reduces the loss contribution of well-classified samples and allows the model to focus on harder examples. The standard cross-entropy loss can be written as:
\begin{equation}\label{cross_entropy}
CE(p_t) = -\text{log}(p_t), \quad
p_t=
\begin{cases}
    p, & \text{if} \ y=1        \\
    1-p, & \text{if} \ y=0
\end{cases}
\end{equation}
The loss contribution from well-classified samples (e.g., easy negatives with $p_t \approx 1$) dominates training, leading to under-representation of hard-to-classify samples, particularly the minority class. This can cause the model to bias heavily toward the majority class.

Focal Loss introduces a modulating factor $(1-p_t)^{\gamma}$ to the cross-entropy loss, which dynamically reduces the loss contribution from well-classified examples ($p_t \to 1$) and focuses more on hard examples ($p_t \to 0$). The Focal Loss is defined as:
\begin{equation}\label{focal_loss}
    FL(p_t) = -\alpha _t (1-p_t)^{\gamma}\text{log}(p_t).
\end{equation}

Quality Focal Loss (QFL) \cite{Li10.5555} is an extension of Focal Loss designed to improve object detection by jointly optimizing classification and localization quality. Unlike standard classification losses that treat class labels as binary values (0 or 1), QFL assigns a continuous quality score (e.g., Intersection over Union (IoU) between the predicted and ground truth bounding boxes) as the supervision target. This helps the model learn confidence-aware classification scores, reducing false positives.

QFL generalizes the standard Focal Loss to support continuous supervision labels instead of discrete ones. It is defined as:
\begin{equation}\label{qfl}
L_{QFL} = -|y-\sigma|^{\beta}[y\text{log}\sigma+(1-y)\text{log}(1-\sigma)],
\end{equation}
where $y \in [0, 1]$ is the soft target representing the IoU score between the predicted and ground truth bounding boxes, $p$ donates the predicted confidence score of the object, and $\beta$ is a modulating parameter that adjusts the weighting of hard examples. Unlike the original Focal Loss, which uses discrete labels ($y \in \{0,1\}$), QFL enables a continuous label space, allowing more nuanced learning of classification confidence based on localization quality.

YOLOv8’s loss function is composed of three key components: complete-IoU (cIoU) \cite{Zheng9523600} for bounding box regression, Distribution Focal Loss (DFL) \cite{Li10.5555} for box refinement, and Binary Cross Entropy (BCE) \cite{Cox1958TheRA} for classification and objectness. In this setup, we replace BCE with QFL to address class imbalance.

\begin{table*}[htbp]
    \centering
    \renewcommand{\arraystretch}{1.2}
    \caption{Comparison with state-of-the-art object detection methods on the proposed thermal dataset for people with mobility restrictions.}
    \label{Comparison with sota object detection methods}

    \begin{tabular*}{\textwidth}{@{\extracolsep{\fill}} 
    p{0.18\linewidth}|cccccccc} 
    \toprule
    Model & Params (M) $\downarrow$ & GFLOPs $\downarrow$ & FPS $\uparrow$ & AP$^{val}$ $\uparrow$
    & AP$^{val}_{50}$ $\uparrow$ & AP$^{val}_{75}$ $\uparrow$ & AP$^{val}_{S}$ $\uparrow$ 
    & AP$^{val}_{L}$ $\uparrow$ \\
    \midrule
    YOLOv11-L & \textbf{25.3}  & \textbf{87.6}  & 98.0 & 87.6 & 93.7 & 92.0 & 71.1 & 88.3 \\
    YOLOv11-X & 56.9  & 196.0 & 51.3 & 87.5 & 94.4 & 90.8 & 73.9 & 88.1 \\
    YOLOv10-L & 25.7  & 126.4 & 85.5 & 87.1 & 93.3 & 91.7 & 68.4 & 87.9 \\
    YOLOv10-X & 31.6  & 169.9 & 53.8 & 86.1 & 91.4 & 90.0 & 69.1 & 86.9 \\
    YOLOv9-C  & 21.1  & 82.7  & 104.2 & 88.4 & 94.7 & 92.8 & 70.0 & 89.1 \\
    YOLOv9-E  & 54.0  & 173.4 & 47.6 & 88.8 & 94.2 & 92.3 & \textbf{75.5} & 89.5 \\
    YOLOv8-L  & 39.4  & 145.2 & 88.5 & 88.1 & 93.7 & 92.3 & 69.6 & 88.9 \\
    YOLOv8-X  & 61.6  & 226.8 & 55.9 & 88.4 & 94.7 & 92.3 & 72.2 & 89.3 \\
    YOLOv7-L  & 36.5  & 103.3 & \textbf{169.5} & 79.5 & 87.6 & 85.0 & 61.9 & 80.1 \\
    YOLOv7-X  & 70.9  & 188.2 & 105.3 & 82.9 & 90.7 & 88.6 & 64.3 & 83.4 \\
    YOLOv6-L  & 109.6 & 387.0 & 59.2 & 83.3 & 89.2 & 87.3 & 71.5 & 83.9 \\
    YOLOv6-X  & 171.0 & 603.4 & 38.2 & 81.7 & 87.8 & 86.0 & 69.3 & 82.3 \\
    RT-DETR (R50) & 42.0 & 125.7 & 60.2 & 86.2 & 92.4 & 90.7 & 69.3 & 87.1 \\
    RT-DETR (R101) & 60.9 & 186.3 & 45.7 & 88.6 & 94.0 & 92.6 & 69.1 & 89.5 \\
    Ours      & 42.1  & 150.1 & 90.1 & \textbf{89.1} & \textbf{95.1} & \textbf{93.0} & 71.5 & \textbf{89.9} \\

   \bottomrule
\end{tabular*}
\end{table*}

\section{Experiments}
\label{Experiments}
\subsection{Comparison with SOTA: Real-Time Object Detection}
We conduct a comprehensive evaluation that includes both qualitative performance assessment and an analysis of computational complexity, measured in floating-point operations per second (FLOPs) and overall parameter count. To facilitate this evaluation, we benchmark state-of-the-art (SOTA) real-time object detection methods using our own thermal dataset TD4PWMR, training the models on its training partition and evaluating them on the test set. The corresponding quantitative results are presented in Table \ref{Comparison with sota object detection methods}, where the $L$ and $X$ models of YOLO detector are evaluated. These models have higher parameter counts, demand greater computational power, and result in longer inference times while achieving superior accuracy. To ensure a fair comparison, we retrain these models using the hyperparameters specified by the original authors. The detectors utilize a commen input size of $640 \times 640$ pixels. The FPS is tested on GeForce RTX 2080 Ti GPU.

We evaluate model performance using COCO-style metrics. AP$^{val}$ serves as the primary metric for object detection, computed by averaging the area under the precision-recall curve across multiple Intersection-over-Union (IoU) thresholds, ranging from 0.5 to 0.95 in increments of 0.05. AP$_{50}^{val}$ and AP$_{75}^{val}$ represent the average precision computed at fixed IoU thresholds of 0.50 and 0.75, respectively. Additionally, AP$_L^{val}$ and AP$_L^{val}$ split the evaluation based on object size with the $96\times96$ pixel threshold, computing the AP for small and large objects, respectively.

The proposed model achieves the highest overall accuracy among all evaluated methods, attaining an AP$^{val}$ of $89.1\%$. It also outperforms all competitors in terms of AP$_{50}^{val}$ ($95.1\%$) and AP$_{75}^{val}$ ($93.0\%$), highlighting its robustness across various IoU thresholds. Furthermore, in detecting large objects, our approach achieves the highest AP$_L^{val}$ ($89.9\%$).  

In addition to its strong detection performance, the proposed model efficiently balances accuracy and computational complexity. With a moderate parameter count (42.1M) and computational cost (150.1 GFLOPs), it remains significantly more efficient than larger models such as YOLOv6-L (171.0M, 603.4 GFLOPs) while achieving superior accuracy. The model also achieves a high inference speed of 90.1 FPS, ranking among the fastest methods evaluated. It surpasses most competitors in inference efficiency, with only YOLOv7-L (169.5 FPS), YOLOv7-X (105.3 FPS), YOLOv9-C (104.2 FPS), and YOLOv11-L (98.0 FPS) achieving higher inference speed.  

Compared to emerging real-time transformer-based models, our proposed approach demonstrates clear advantages. It consistently outperforms RT-DETR (R50) across all key evaluation metrics, achieving a higher AP$^{val}$ ($89.1\%$ vs. $86.2\%$) and AP$^{val}_{50}$ ($95.1\%$ vs. $92.4\%$), while delivering significantly improved inference speed (90.1 FPS vs. 60.2 FPS). Similarly, in comparison to RT-DETR (R101), our method attains a slight performance gain in AP$^{val}$ ($89.1\%$ vs. $88.6\%$) and AP$^{val}_{50}$ ($95.1\%$ vs. $94.0\%$), while nearly doubling the inference speed (90.1 FPS vs. 45.7 FPS).

Based on both qualitative and quantitative assessments, we conclude that the proposed object detection model outperforms SOTA performance while requiring fewer computational resources, achieving a lower parameter count and reduced computational complexity.

\begin{table*}[htbp]
    \centering
    \renewcommand{\arraystretch}{1.2}
    \caption{Ablation study on the validation set of the proposed thermal dataset for people with
mobility restrictions.}
    \label{Ablation study}
    \begin{tabular*}{\textwidth}{@{\extracolsep{\fill}} l|cccc|ccccccc}
    \toprule
        Method & Triplet-Attention & SPD-Conv & SPPFCSPC & QFL & AP$^{val}$ $\uparrow$ 
        & AP$_{50}^{val}$ $\uparrow$ & AP$_{75}^{val}$ $\uparrow$ & AP$_S^{val}$ $\uparrow$ & AP$_L^{val}$ $\uparrow$ 
        &  FPS $\uparrow$ & \\
        \midrule
        Baseline         &        &         &         &         & 88.1 & 93.7 & 92.3   
                         & 69.6 & 88.9 & 88.5   \\
        YOLO-Thermalv1   & \cmark &         &         &         & 88.4 & 94.7 & 92.0   
                         & \textbf{72.1} & 89.1 & 87.7 \\
        YOLO-Thermalv2   & \cmark & \cmark  &         &         & 88.6 & 94.2 & 92.9
                         & 71.1 & 89.4 & \textbf{90.9} \\
        YOLO-Thermalv3   & \cmark & \cmark  & \cmark  &         & 88.7 & 94.7 & 92.4
                         & 71.0 & 89.4 & 90.1 \\
        YOLO-Thermalv4   & \cmark & \cmark  & \cmark  & \cmark  & \textbf{89.1} & \textbf{95.1} & \textbf{93.0}    & 71.5 & \textbf{89.9} & 90.1\\        
        \bottomrule
    \end{tabular*}
\end{table*}

\begin{table*}[htbp]
\centering
\caption{Ablation study on the number of consecutive frames in the multi-frame validation approach.}
\label{N_wasted_time_accuracy}
\begin{tabular*}{\textwidth}{@{\extracolsep{\fill}}c|cccccc}
\toprule
\multirow{2}{*}{Parameter $N$} & 
\multirow{2}{*}{Success rate (\%) $\uparrow$} & 
\multirow{2}{*}{Latency ($s$) $\downarrow$} & 
\multirow{2}{*}{Real success rate ($\%$) $\uparrow$} & 
\multicolumn{3}{c}{Average extended green time ($s$) for people with} \\
\cmidrule(lr){5-7}
& & & & walking impairment & visual impairment & mobility burden \\
\midrule
1 & 13.4 & 0.8 & 32.3 & 1.6 & 2.2 & 0.8 \\
2 & 77.2 & 1.2 & 95.4 & 2.9 & 3.9 & 1.3 \\
3 & 92.7 & 1.6 & 96.6 & 3.1 & 4.2 & 1.6 \\
4 & 94.6 & 1.9 & 96.9 & 3.3 & 4.3 & 1.7 \\
5 & 94.8 & 2.3 & 97.0 & 3.5 & 4.5 & 1.9 \\
\bottomrule
\end{tabular*}
\end{table*}

\subsection{Ablation Study}

Table \ref{Ablation study} presents an ablation study conducted on the validation set of the proposed thermal dataset for people with mobility restrictions. The study evaluates the performance of different configurations of the YOLO-Thermal model, starting from YOLOv8, which serves as the baseline, and progressively adding different components.
\subsubsection{Triplet-Attention} Adding Triplet-Attention improves feature representation, leading to a noticeable increase in AP$^{val}$ from $88.1\%$ to $88.4\%$ and AP$_{50}^{val}$ from $93.7\%$ to $94.7\%$. Small object detection (AP$_S^{val}$) also improves from $69.6\%$ to $72.1\%$, indicating better feature extraction for fine details. However, the FPS drops slightly to 87.7, suggesting a minor computational overhead.

\subsubsection{SPD-Conv}
Introducing SPD-Conv further enhances performance, boosting AP$^{val}$ to $88.6\%$ and AP$_{75}^{val}$ to $92.9\%$, indicating improved precision at higher IoU thresholds. Additionally, the FPS is slightly improved to 90.9, demonstrating a more efficient inference process. Notably, SPD-Conv does not improve the detection accuracy of small objects (AP$_S^{val}$) and even degrades performance after integration into the model, contradicting the findings of the original paper. Nevertheless, it has been shown to enhance the overall detection accuracy of thermal images, aligning with the finding that it can improve detection in low-resolution images.

\subsubsection{SPPFCSPC}
The addition of SPPFCSPC continues the trend of improvement, improving AP$^{val}$ to $88.7\%$ and restoring AP$_{50}^{val}$ to $94.7\%$. However, the FPS drops slightly to 90.1, reflecting an increased computational cost.

\subsubsection{QFL}
Finally, the incorporation of QFL significantly enhances the detection of hard-to-classify objects, pushing AP$^{val}$ to $89.1\%$ — the highest in this study. The improvement is evident across most metrics, with AP$_{50}^{val}$ increasing to $95.1\%$ and AP$_L^{val}$ reaching $89.9\%$.

We deployed the proposed model on a local edge computing device (NVIDIA Jetson Orin), achieving an inference time of 40.4 milliseconds per frame. Given the parallel processing of eight camera streams, the effective interval between consecutive frames from the same stream is 363.4 milliseconds, affecting the detection frequency and influencing the optimal configuration of the number of consecutive frames, denoted by parameter $N$, in the multi-frame validation approach. 

To analyze the impact of $N$, we conducted an ablation study summarized in Table~\ref{N_wasted_time_accuracy}. The table reports four key metrics: (1) success rate, defined as the proportion of events where the system successfully extends the green phase with sufficient addtional time, under the assumption that the extension is terminated immediately at the start of the multi-frame validation approach, once no people with mobility restrictions are detected remaining in the crosswalk; (2) latency caused by the multi-frame validation approach and the integer-valued constraints of the traffic light controller; (3) real success rate, which is introduced to reflect the system's real performance under deployment conditions by adjusting the nominal success rate by accounting for latency; and (4) the average extended green time allocated to people with walking impairment, visual impairment, and mobility burden, respectively.

For $N=1$, wasted time is minimal ($0.8\,s$), but the success rate remains low (13.4 $\%$) and the real success rate is only $32.3\%$, indicating frequent failures to adequately extend the green phase. At $N=2$, performance improves significantly. The success rate rises to $77.2\%$, and the real success rate reaches $95.4\%$, effectively satisfying the design criterion of supporting at least $95\%$ of people with mobility restrictions. The latency increases moderately to $1.2\,s$, which remains acceptable with respect to traffic efficiency. The average extended crossing times are $2.9\,s$ for people with walking impairments, $3.9\,s$ for people with visual impairments, and $1.3\,s$ for people with mobility burdens. While increasing $N$ to 3 marginally improves the success rate ($92.7\%$) and the real success rate ($96.6\%$), it introduces higher latency ($1.6\,s$). Further increases to $N=4$ and $N=5$ result in diminishing returns, with the real success rate plateauing at approximately $97.0\%$, while latency rises significantly to $1.9\,s$ and $2.3\,s$, respectively. 

Notably, once $N>2$, the observed increases in average extended green time are primarily driven by latency rather than substantive improvements in the reliability of the multi-frame validation approach. This outcome is undesirable, as it degrades traffic efficiency without yielding proportional gains in the real success rate, thereby offering no additional benefits in terms of safety and accessibility for pedestrians with mobility impairments. Furthermore, the latency introduced by the integer-valued constraints of the traffic light controller remains approximately constant at $0.5\,s$ across different values of $N$. This indicates that the dominant source of increased latency at higher $N$ originates from the delays intrinsic to the the multi-frame validation process, rather than from the integer-valued constraints of the traffic light controller.

Therefore, setting $N$ to 2 provides a balanced tradeoff between ensuring safety and accessibility of pedestrians with mobility restrictions and minimizing unnecessary latency, considering the constraints of our edge computing system and multi-stream processing setup.

\begin{figure*}[htbp]
    \centering
    \captionsetup[subfloat]{labelformat=empty, justification=centering, font=scriptsize}

    \subfloat[]{%
        \includegraphics[width=0.195\textwidth]{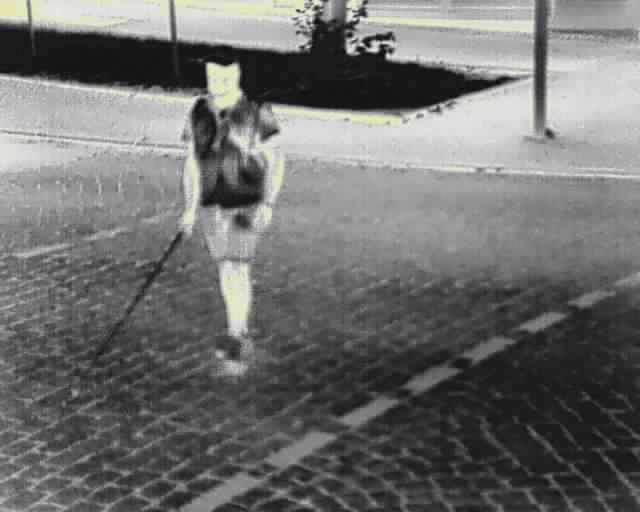}%
    }
    \hfill
    \subfloat[]{%
        \includegraphics[width=0.195\textwidth]{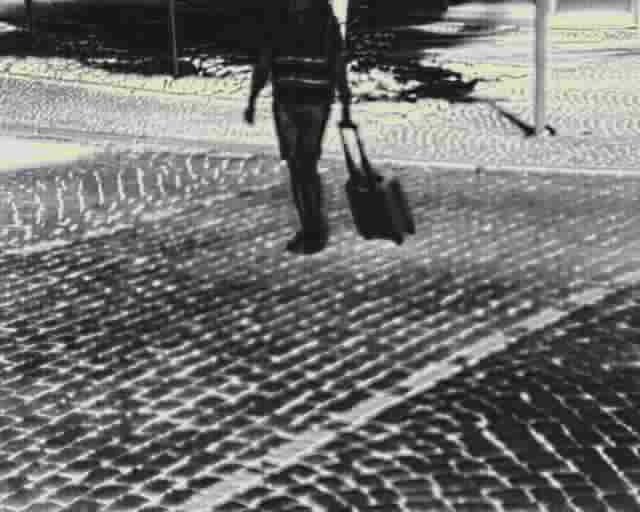}%
    }
    \hfill
    \subfloat[]{%
        \includegraphics[width=0.195\textwidth]{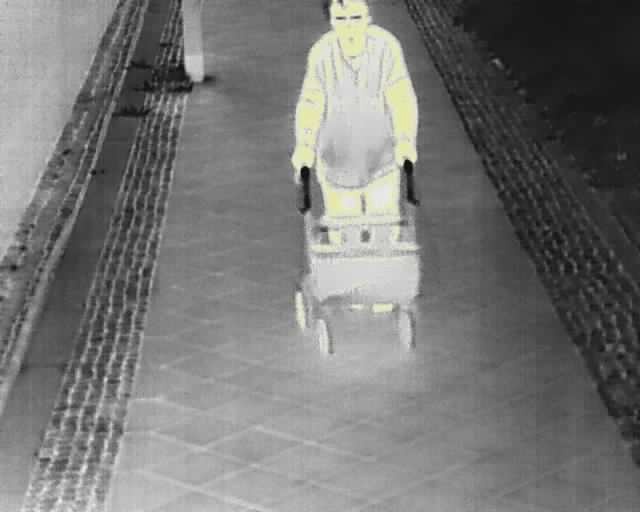}%
    }
    \hfill
    \subfloat[]{%
        \includegraphics[width=0.195\textwidth]{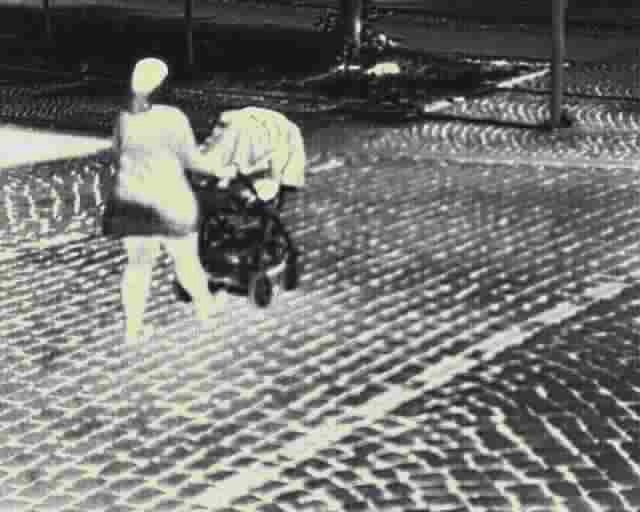}%
    }
    \hfill
    \subfloat[]{%
        \includegraphics[width=0.195\textwidth]{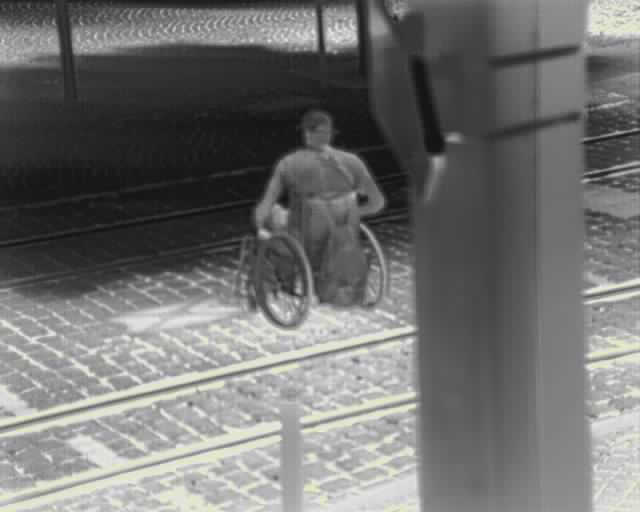}%
    }

    \vspace{-0.76cm} 

    \subfloat[]{%
        \includegraphics[width=0.195\textwidth]{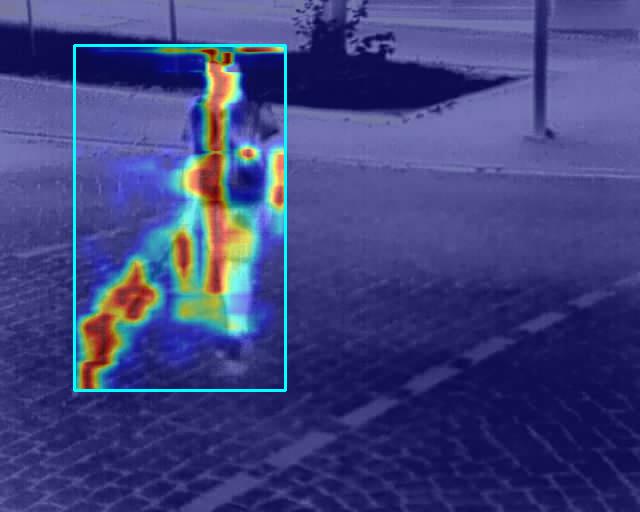}%
    }
    \hfill
    \subfloat[]{%
        \includegraphics[width=0.195\textwidth]{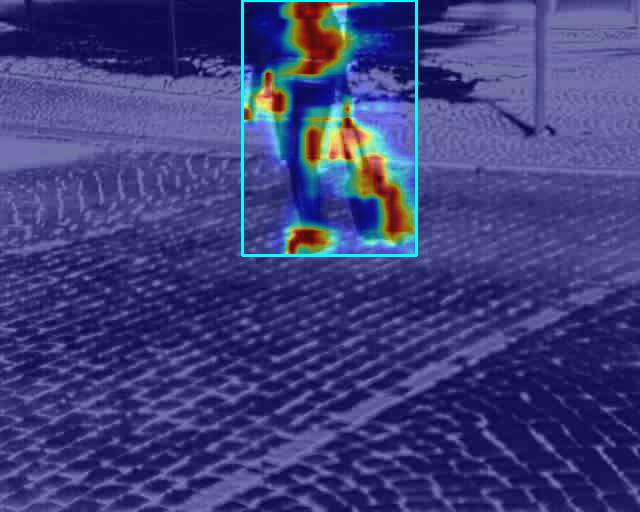}%
    }
    \hfill
    \subfloat[]{%
        \includegraphics[width=0.195\textwidth]{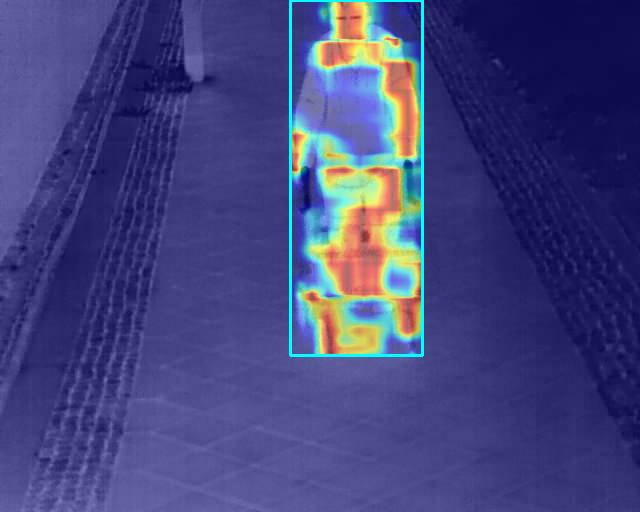}%
    }
    \hfill
    \subfloat[]{%
        \includegraphics[width=0.195\textwidth]{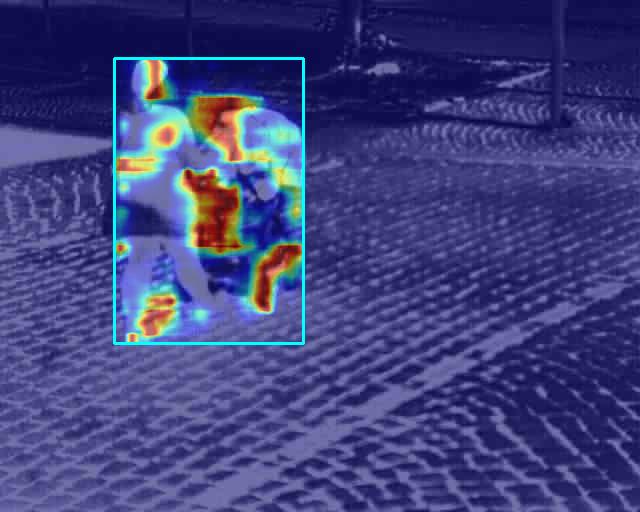}%
    }
    \hfill
    \subfloat[]{%
        \includegraphics[width=0.195\textwidth]{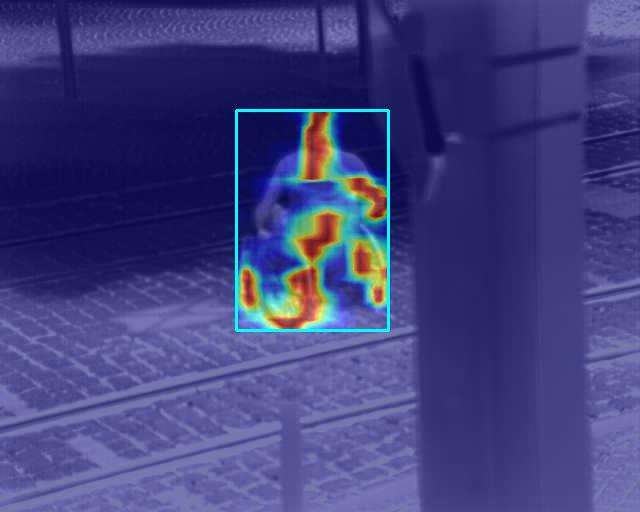}%
    }

    \caption{Visualization of raw images (top) and corresponding LayerCAM heatmaps (bottom). Warmer colors indicate regions of higher activation.}
    \label{Visualization}
\end{figure*}

\subsection{Visualization}
To further investigate the discriminative capacity of the learned features, we conduct a retrieval-based visualization experiment using LayerCAM~\cite{Jiang9462463}, as shown in Fig.~\ref{Visualization}. LayerCAM is a gradient-based class activation mapping method that leverages gradient information from intermediate convolutional layers to generate spatially-aware heatmaps, highlighting the regions most influential to the model’s prediction.

We visualize model attention in scenarios involving individuals using a mobility aid or with mobility burden. The top row of Fig.~\ref{Visualization} shows the input thermal images, while the bottom row presents the corresponding LayerCAM heatmaps. The highlighted activations consistently localize around both the person and the associated object, particularly in the region that connects the two. This indicates that the model captures not only individual features but also the relational cues between the human subject and the associated object.

This observation supports our hypothesis that the relative position between the person and the associated object is critical for accurate detection. Moreover, the consistent activation patterns validate our annotation strategy, which treats the individual and the associated object as a single instance. This choice of annotation strategy effectively enables the detection model to better capture the semantic and spatial coherence of human-object interaction, thereby improving both detection accuracy and interpretability.

\section{Conclusion}
\label{Conclusion}
This paper presents a novel thermal detector-based smart traffic light designed to enhance accessibility and safety for pedestrians with mobility restrictions. We introduce TD4PWMR, a specialized thermal dataset that captures diverse pedestrian scenarios across various environmental conditions, addressing critical gaps in existing datasets. To improve detection accuracy in thermal imaging, we propose YOLO-Thermal, an optimized object detection framework incorporating advanced feature extraction techniques such as Triplet-Attention, SPD-Conv, and SPPFCSPC. Experimental results demonstrate that YOLO-Thermal outperforms state-of-the-art models in both accuracy and efficiency. Finally, we implement an adaptive traffic light control strategy that dynamically adjusts green light durations and enhances auditory guidance based on real-time pedestrian detection. The proposed system significantly improves intersection accessibility while maintaining efficient traffic flow. 

In future work, we aim to further enhance the system's real-time performance by integrating model compression techniques such as knowledge distillation and pruning. These approaches are expected to reduce computational overhead and accelerate inference, thereby mitigating delays introduced by the current multi-frame validation strategy. By minimizing unnecessary extensions of green light durations when no people with mobility restrictions still remain in the crosswalk, this optimization will contribute to a more responsive and efficient traffic control system that balances accessibility with urban mobility needs.

\section{Acknowledgment}
This work was funded by the Federal Ministry of Digital and Transport of Germany – 45AVF3005A-E.


%





\ifCLASSOPTIONcaptionsoff
  \newpage
\fi





\bibliographystyle{IEEEtran}
\bibliography{IEEEabrv,Bibliography}
%

\begin{IEEEbiography}[{\includegraphics[width=1in,height=1.25in,clip,keepaspectratio]{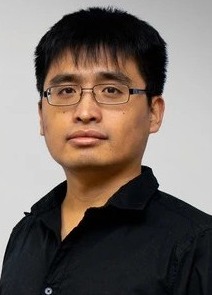}}]{Xiao Ni}
received the B.Eng. degree from Tongji University and is currently pursuing the Ph.D. degree from the University of Münster. His research interests include thermal image processing and object tracking.
\end{IEEEbiography}
\begin{IEEEbiography}[{\includegraphics[width=1in,height=1.25in,clip,keepaspectratio]{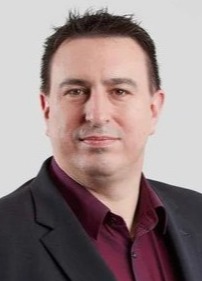}}]{Carsten Kühnel}
received his diploma in Industrial Engineering (Civil Engineering) from the Technical University of Darmstadt, Germany in 2004, followed by the Ph.D. degree in 2011 from the University of Kassel, Germany. He worked as the team leader for Innovative Technologies and Cooperative Systems at the Traffic Center Hessen and was responsible for the deployment of C-ITS in Germany within the C-ITS-Corridor project amongst others. Since 2017, he has been a Professor for Intelligent Transportation Systems at the University of Applied Sciences Erfurt (FHE), Germany. Accompanied by two colleagues he leads the Institute for Transport and Spatial Planning at the FHE. He is a member of different working groups of the German Road and Transportation Research Association (FGSV), for example as the vice chair of the working group 3.1 Telematics.
\end{IEEEbiography}
\begin{IEEEbiography}[{\includegraphics[width=1in,height=1.25in,clip,keepaspectratio]{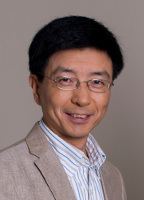}}]{Xiaoyi Jiang}
received the bachelor’s degree from Peking University, Beijing, China, and the Ph.D. and Venia Docendi (Habilitation) degrees from the University of Bern, Bern, Switzerland, all in Computer Science. He was an Associate Professor with the Technical University of Berlin, Berlin, Germany. Since 2002, he has been a Full Professor of Computer Science with the University of Münster, Münster, Germany, where he was the Dean of the Faculty of Mathematics and Computer Science (2016--2023). His current
research interests include pattern recognition, image analysis, and biomedical imaging. Dr. Jiang is an Editor-in-Chief of International Journal of Pattern Recognition and Artificial Intelligence.
He also serves on the Advisory Board and Editorial Board of several journals, including International Journal of Neural Systems and Journal of Big Data.
Previously, he has been Associate Editor for
IEEE Trans. on Systems, Man, and Cybernetics - Part B /  IEEE Trans. on Cybernetics, IEEE Trans. on Medical Imaging, and Pattern Recognition. He is  a Senior Member of IEEE and a Fellow of IAPR.
\end{IEEEbiography}





\vfill


\end{document}